# Title: Deep Learning for Localization in the Lung

**Authors:** Jake Sganga[1*], David Eng[2], Chauncey Graetzel[3], David B. Camarillo[1]

**Affiliations:**

[1]Stanford University, Department of Bioengineering.

[2]Stanford University, Department of Computer Science.

[3]Auris Health, Inc.

*Correspondence to: sganga@stanford.edu

**Abstract:** Lung cancer is the leading cause of cancer-related death worldwide, and early diagnosis is critical to improving patient outcomes. To diagnose cancer, a highly trained pulmonologist must navigate a flexible bronchoscope deep into the branched structure of the lung for biopsy. The biopsy fails to sample the target tissue in 26-33% of cases largely because of poor registration with the preoperative CT map. We developed two deep learning approaches to localize the bronchoscope in the preoperative CT map in real time and tested the algorithms across 13 trajectories in a lung phantom and 68 trajectories in 11 human cadaver lungs. In the lung phantom, we observe performance reaching 95% precision and recall of visible airways and 3 mm average position error. On a successful cadaver lung sequence, the algorithms trained on simulation alone achieved 77%-94% precision and recall of visible airways and 4-6 mm average position error. We also compare the effect of GAN-stylizing images and we look at aggregate statistics over the entire set of trajectories.

**One Sentence Summary:** Neural networks trained on simulated data can track a bronchoscope's movement through a plastic lung phantom and a human cadaver lung.

**Main Text:**

Diagnosing lung cancer, the leading cause of cancer-related death world-wide, at an early stage significantly improves patient outcomes (*1*). Physicians biopsy potentially cancerous nodules in the lung through bronchoscopies, where they manually drive long, flexible bronchoscopes through the patient's airways, shown in Fig. 1. This minimally invasive approach is preferred when the nodule is accessible, given the lower complication rates (2.2% vs 20.5%) compared to thoracoscopic surgery or transthoracic needle biopsy (*2*, *3*).

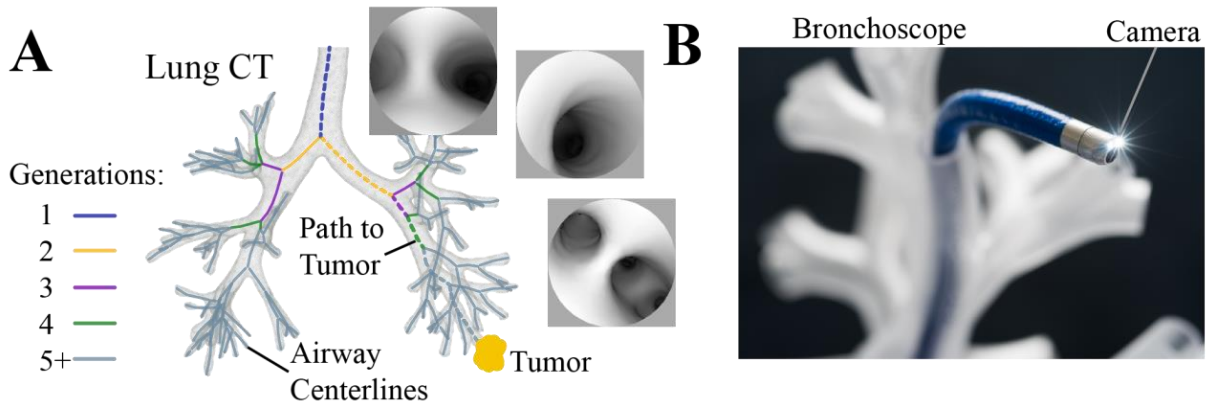

**Fig. 1**. In A, a path through a preoperative lung CT is shown toward the site of a potential tumor. Images shown were rendered along the path to demonstrate what the physician would see as they move through the generations of the lung. The branching structure of the lung is represented in a lower dimensional skeletal tree based on the airway centerlines and the junctions between them, usually bifurcations. Planned trajectories through the lung will consist of series of airway centerlines. In B, a robotic bronchoscope (Auris Health, Inc.) is shown and the distal monocular camera used for visualizing the airways is highlighted.

Before the bronchoscopy, the physician selects biopsy targets in the lung's computed tomography (CT) scan. To reach these targets, the physician maps the sensor feedback from the device (2D image) to the CT frame (3D map). This process is called localization (*4*). With accurate localization information, a physician can select the correct airways leading to the biopsy targets.

Recently, robotic endoscopy systems have been developed to further aid the physician in reaching the target (*5*). The actuators enable the endoscope to take more complex shapes that better conform to anatomy and that can be held over extended periods of time without inducing user fatigue. The robot telemetry aids in the localization process as the commanded motion can be used to update the estimated position. The user interface can simplify driving by augmenting the raw sensor data with the patient CT. The systems offer various levels of control guidance and assistance. Longer term, if the localization were extremely precise, a closed-loop control system could drive the bronchoscope without human intervention.

Other sensing modalities can assist in localization, including a distal electromagnetic (EM) position sensor. After registration to the preoperative computed tomography (CT) of the patient's chest, the position sensor can enable GPS-like directions on the road map to the target site (*6*). Despite the promise of navigated bronchoscopies, institutions vary in the diagnostic yield of these procedures, ranging from 67-74% (*7*, *8*). Additionally, techniques like fluoroscopy, radial ultrasound probes, and alternative endoscopic sensors (Raman spectroscopy, confocal, etc.) have been developed to further improve diagnostic yield (*9*). We choose to focus on advancing image-based approaches since cameras are the cheapest and most prevalent sensor for bronchoscopies, and information from the camera can be integrated with other modalities (*10*).

An image-based system must satisfy two requirements to aid decision-making: 1. it must provide accurate localization and 2. it must be real-time. Several groups have compared the images from the bronchoscope to simulated images rendered from the estimated location of the bronchoscope in CT frame; however, these methods register images inefficiently at around 1-2

Hz (*8*, *11*, *12*). Tracking features using methods like SIFT and ORBSLAM have been used, but the airways have insufficient features and tracked features often drop out (*13*, *14*). Anatomical landmarks have been tracked, like bifurcations (*15*), lumen centers (*16*), centerline paths (*17*), or similar image regions (*18*), but these approaches may not operate in real-time and tend to struggle with image artifacts since they make assumptions about the airway geometries. Merritt et al. describes a real-time localization approach and reports continuous tracking, but it relies on high-quality rendering and a dense collection of reference images (*19*). Finally, nearly all of the previous work use different datasets to evaluate their models, which makes comparing approaches challenging.

Because of the difficulties traditional computer vision techniques face in this task, we decided to explore a deep learning approach. Using convolutional neural networks (CNN) to estimate the position and orientation of objects has been shown in many contexts, including for human posture and objects in a hand (*20*, *21*). The KITTI dataset is a large, high quality dataset to improve visual-based localization methods for autonomous cars (*22*), and the top performing algorithms use variations of CNNs to process the visual information. In the lung domain, Visentini-Scarzanella et al. used a CNN to estimate the depth map of 2D images in a lung phantom, which could then be registered to the 3D map, but localization is not reported (*14*). In our previous work, we used a CNN to localize a bronchoscope in real-time by predicting the offset between the camera image and a rendering of the expected position (*23*). This approach showed 1.4 mm accuracy on a lung phantom sequence, but it fails to track other sequences.

In this work, we contribute two image-based deep-learning approaches, called BifurcationNet and AirwayNet, that localize a bronchoscope in the lung CT frame accurately and in real-time. We evaluate both approaches on two datasets: one recorded from a silicone lung phantom and one recorded from trajectories in eleven human cadaver lungs. Both approaches demonstrate continuous, real-time tracking on sequences in the lung phantom and human cadaver lungs after training on simulated images alone. Furthermore, we evaluate the performance of the networks on holdout cadaver lungs when trained on a range from one other lung to ten other lungs. Despite the algorithms only inconsistently localizing on cadaver lung sequences, we demonstrate the proof of concept for our algorithms in a challenging, clinically-relevant dataset.

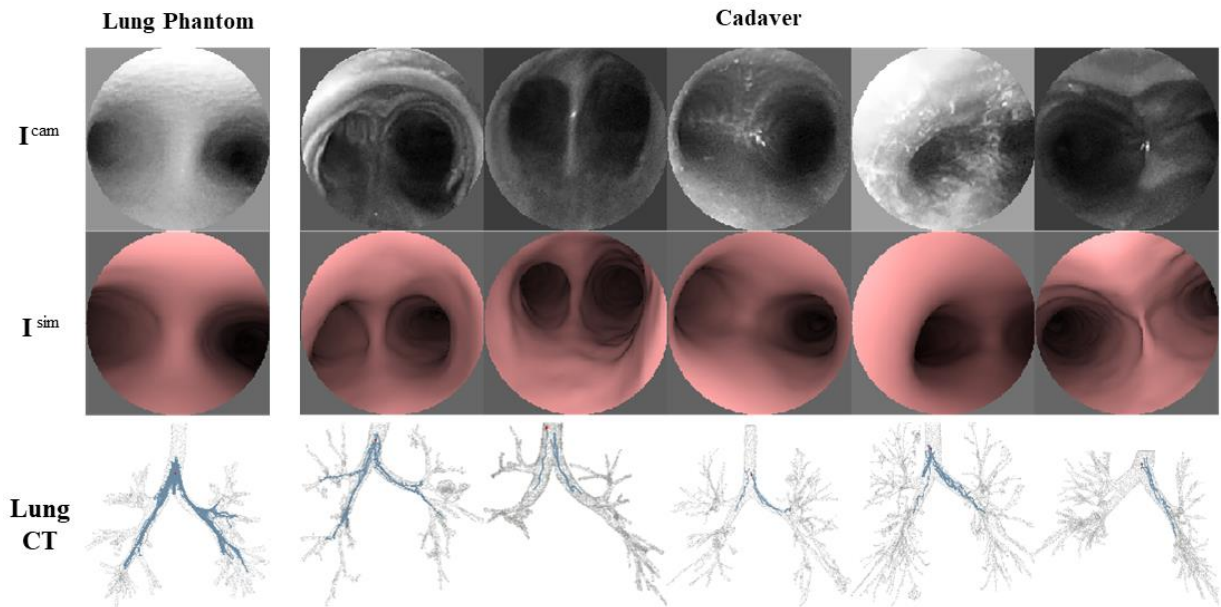

**Fig. 2**. Example camera images, $I^{cam}$, in the main carina from the lung phantom and five of the eleven cadavers are shown on the top row. Below each image is the simulated rendering of that location in the respective lung CT scans, $I^{sim}$. The full lung CT scan and the tested paths through each lung is shown in the bottom row.

**Table 1**. Notation, modified from (*23*)

| | **Image Styles** |
|---|---|
| $I^{cam}$ | Image taken by a bronchoscope, applies to phantom and cadaver lungs |
| $I^{sim}$ | Image rendered by OpenGL using the lung CT |
| $I^{rsim}$ | Image rendered by OpenGL using the lung CT with varied rendering parameters (*24*) |
| $I^{gan}$ | Image taken by a bronchoscope, then passed through a generator of a trained GAN (generative adversarial network) to appear styled like $I^{sim}$ |
| | **Image Sources** |
| $I_{x \in L_{i=j}}$ | Image set taken from a set of locations, $x$, in lung, $L_i$, which is the same as the holdout lung, $L_j$ |
| $I_{x \in L_{i \neq j}}$ | Image set taken from a set of locations, $x$, in lung, $L_i$, which is different from the holdout lung, $L_j$ |
| $I_{x \in L_{i \neq j} \times n}$ | Image set taken from a set of locations, $x$, in a set of $n$ lungs, which are all different from the holdout lung, $L_j$ |
| | **Tracking Error** |
| $e_p$ | Position error (mm), defined as $e_p$ in (*19*) |
| $e_d$ | Direction angle error between pointing vectors, $p_z$, of the two views (°), defined as $e_d$ in (*19*) |
| $e_r$ | Roll angle error between the $p_x$ axis after $e_d$ was corrected for between views (°), defined as $e_r$ in (*19*) |

## Results

BifurcationNet and AirwayNet were each tested on two datasets—one recorded in a lung phantom (Koken Inc.) and one recorded in 11 human cadaver lungs. We manually registered every image in the datasets to a position and orientation in the corresponding preoperative lung CT scan. The sequences of images used to evaluate the models were selected to start in the trachea and move toward a target airway without significantly obstructed views.

For each dataset, the tracking performance is measured along several dimensions to provide a detailed view of the how the localization would relate to driving decisions. Since successful navigation critically depends on identifying the parent and child airways at a bifurcation, the results emphasize the F1 score, which is the harmonic mean of precision and recall, on visible airways at bifurcations. The F1 score is broken down by skeletal generation to show how performance is affected as the bronchoscope moves deeper into the branched structure. The mean distance in position, direction, and roll between the labeled point, $x_t$, and the estimate, $\hat{x}_t$, is shown for when a bifurcation is visible and is correctly labeled.

On a laptop PC with a 2.70 GHz CPU, the tracking loop ran at an average of 41 Hz for BifurcationNet and 52 Hz for AirwayNet, while the bronchoscope receives images at a rate of 25-30 Hz.

*Lung Phantom Dataset* - 13 test sequences that cover several generations in the left and right lung were chosen to evaluate the performance of BifurcationNet and AirwayNet. For this test, both algorithms were trained only on domain randomized simulated data, $I^{rsim}$. The results from these trials is shown in Fig. 3. Each dot in Fig. 3 represents the average statistic for a single sequence and the bars show the mean across all sequences, weighted by the number of frames in each sequence. AirwayNet consistently tracks the bronchoscope until the end of the sequence, as seen by the high F1 score through generation 5. BifurcationNet on the other hand, begins to show inconsistent results in generation 3. Despite AirwayNet outperforming BifurcationNet from generation 3 onward ($p < 0.05$), BifurcationNet shows lower average localization errors $e_p$, $e_d$ on correctly labeled frames. The individual sequences show that BifurcationNet can have a bimodal distribution of performance with some sequences successfully reaching later generations and others that cannot recover from entering an incorrect airway.

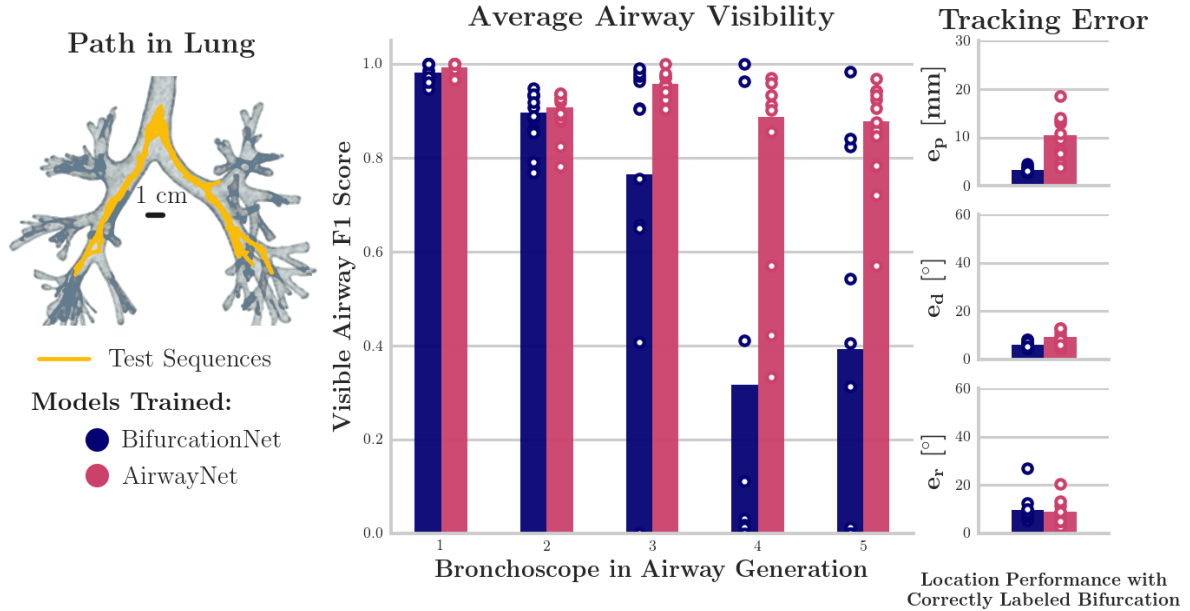

**Fig. 3**. BifurcationNet and AirwayNet, both trained on $I^{rsim}_{x \in L_{i=j}}$, are shown in 13 independent tracking tasks in a lung phantom. Left, the paths of the registered sequences are shown on the lung CT. Middle, the F1 score in classifying airways is shown for each skeletal generation the bronchoscope entered. Right, the tracking analysis shows the error in position, direction and roll for the frames when the airways of a bifurcation were correctly labeled by the algorithm. In the middle and right panels, each dot represents the result from a single sequence and the bar height represents the average across all 13 sequences. AirwayNet outperforms BifurcationNet on the F1 score for generation 3 (p<0.05), and generations 4-5 (p<0.0001).

*Cadaver Lung Dataset* - On the dataset of 68 sequences from 11 cadaver lungs, we evaluated the performance of the algorithms on a holdout cadaver lung as a function of training set domain. Models were trained with the images from one lung, five lungs and ten lungs to see how performance would change with broader training data, denoted $I^{cam}_{x \in L_{i \neq j} \times 1, 5, 10}$, respectively. We also compared these models to models trained on randomized simulated images generated from the holdout lung's preoperative CT scan, $I_{x \in L_{i=j}}^{rsim}$. Finally, we tested the models trained on randomized simulated images against images from the holdout lung that were first passed through a GAN trained on five other lungs to stylize the camera images like the simulated images, $I_{x \in L_{i=j}}^{rsim}$ on $I_{x \in L_j}^{gan}$. This process was repeated for AirwayNet and BifurcationNet, resulting in 92 CNN models and 2 GAN models trained.

Figure 4 shows the aggregate of the sequences with each of the 11 cadaver lungs used as the holdout lung. The bars show the average across all cadaver lungs, weighted by sequence frame length. The error bars show min and max performance of each lung average. For both AirwayNet and BifurcationNet, increasing the number of lungs in the training set improved the average performance on generations 1 and 2 and on the translation error metric, $e_p$. The algorithms' performance further improved when trained on randomized simulated images from holdout lung, $I_{x \in L_{i=j}}^{rsim}$. This trend held for every measured generation for AirwayNet and through generation 4 for BifurcationNet. The tracking errors tended to improve as well. When the networks trained on simulated data were evaluated on GAN-stylized images, $I_{x \in L_j}^{gan}$, the results were mixed, helping the algorithms on some generations, but not consistently improving performance.

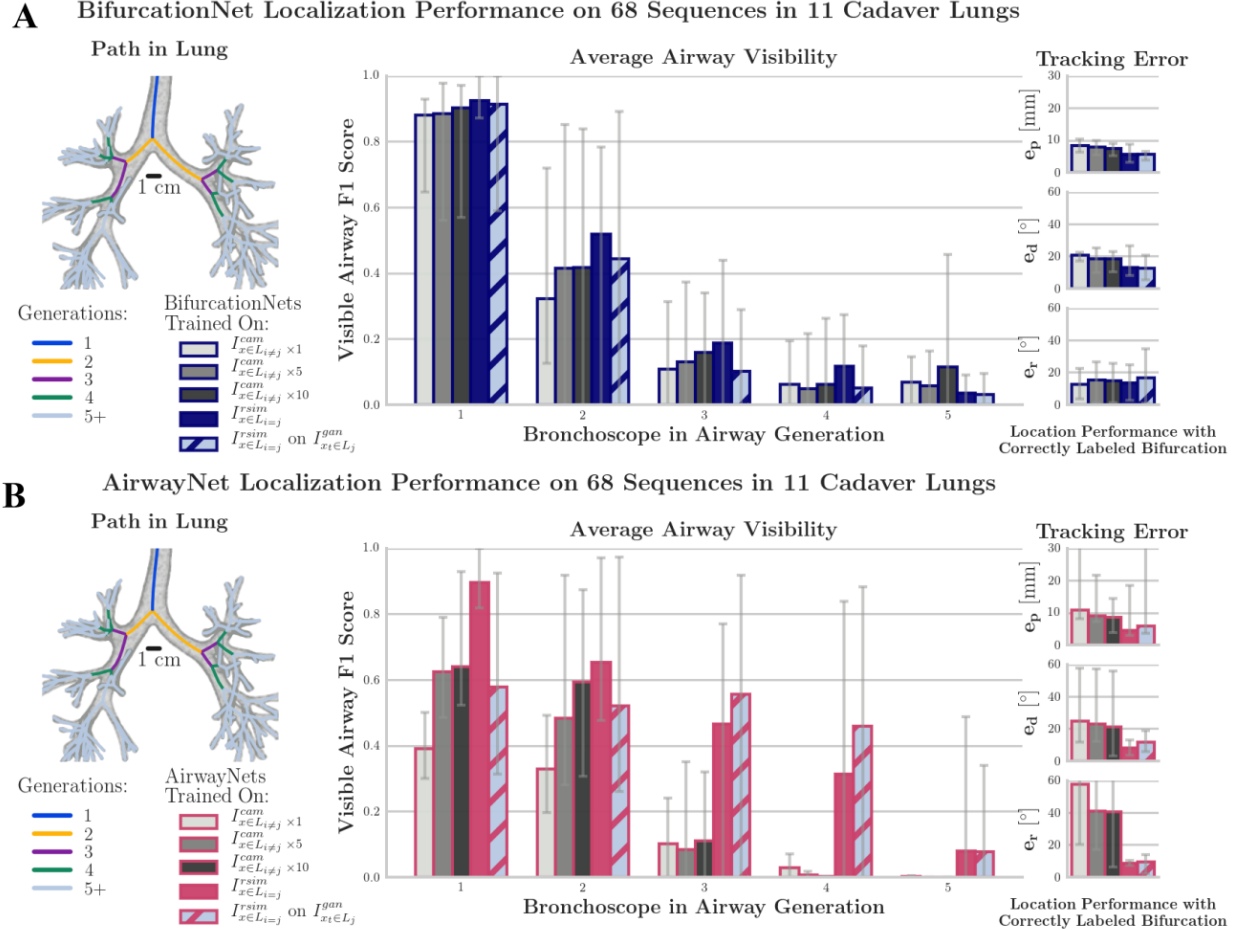

**Fig. 4**. The performance for BifurcationNet, A, and AirwayNet, B, are shown across 68 sequences in 11 cadaver lungs. Each algorithm was evaluated under 4 training conditions -- training on $I_{x \in L_{i \neq j} \times 1}^{cam}$ from 1 lung other than the holdout lung, $I_{x \in L_{i \neq j} \times 5}^{cam}$ from 5 other lungs, $I_{x \in L_{i \neq j} \times 10}^{cam}$ from 10 other lungs, $I_{x \in L_{i = j}}^{rsim}$ in the holdout lung. The model trained on $I_{x \in L_{i = j}}^{rsim}$ in the holdout lung was also evaluated on images after they were passed through a GAN, $I_{x \in L_{j}}^{gan}$, trained on $I_{x \in L_{i \neq j} \times 5}^{cam}$ and $I_{x \in L_{i \neq j} \times 5}^{sim}$ from 5 lungs other than the holdout. The error bars for each result represent the minimum and maximum performance average performance for a holdout lung. Like Fig. 3, the middle panel shows the F1 score in classifying airways is shown for each generation the bronchoscope entered and the right panel shows the tracking analysis when a bifurcation was correctly labeled by the algorithm. AirwayNet outperforms BifurcationNet on the average F1 Score beyond the first generation (p<0.05).

   As an example of a successful cadaver lung sequence for both algorithms, Fig. 5 highlights a sequence that both algorithms localized with low error. In Fig. 5, both algorithms were trained only on randomized simulated data from the holdout lung, $I_{x \in L_{i=j}}^{rsim}$. They track the bronchoscope into the fifth generation, reaching within 4 and 7 mm of the final bronchoscope position for AirwayNet and BifurcationNet, respectively. AirwayNet outperforms BifurcationNet from generation 3 onward and has lower average localization errors $e_p$, $e_d$ on correctly labeled frames.

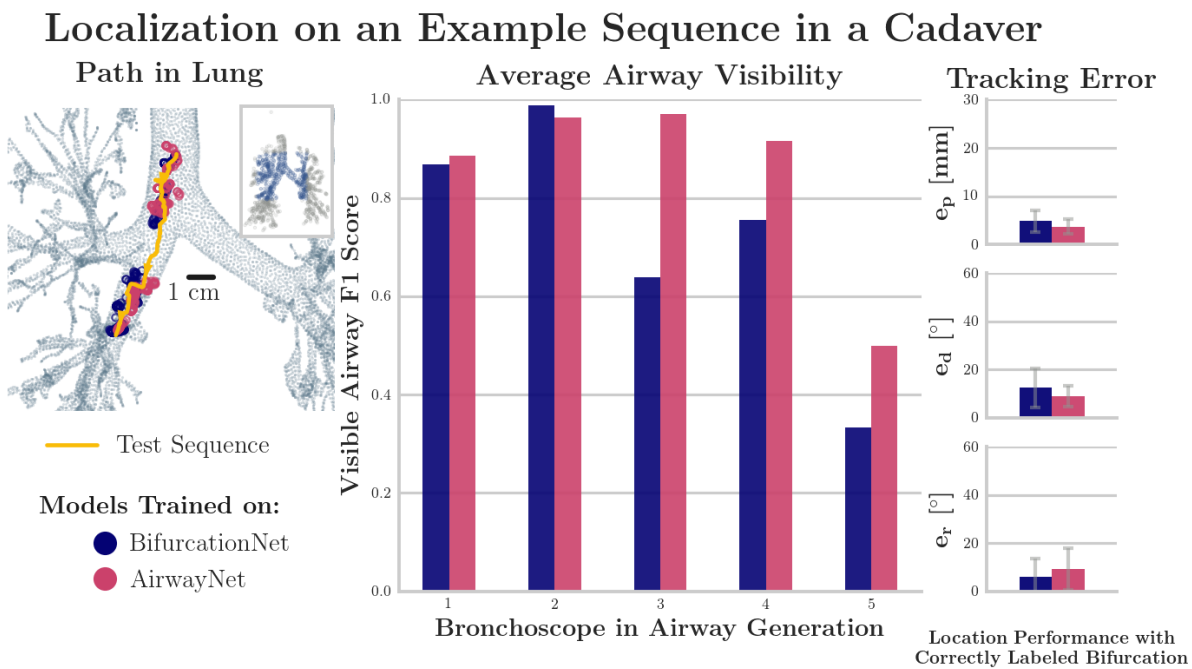

**Fig. 5**. BifurcationNet and AirwayNet, both trained on $I_{x \in L_{i=j}}^{rsim}$, are shown in a tracking task in a human cadaver. Left, the path and estimated positions are shown on the lung CT. Middle, the F1 score in classifying airways is shown for each generation the bronchoscope entered. Right, the tracking analysis shows the error and standard deviation in position, direction and roll for the frames when a bifurcation was correctly labeled by the algorithm.

**Discussion**

This work demonstrated the unexpected result that training deep learning models from simulated images alone can enable real-time, accurate tracking of a bronchoscope in both a lung phantom and human cadaver lungs. The example cadaver lung sequence, shown in Fig. 5, highlights successful localization in a sequence for both BifurcationNet and AirwayNet. This performance is similar to the networks' performance in the lung phantom, demonstrating a proof of concept that the approach used for the lung phantom can apply to human cadaver lungs. However, the results from all 11 cadaver lungs show the consistency in the algorithms' performance can be improved, especially for BifurcationNet, which has variable outcomes even among sequences in the lung phantom.

In the lung phantom dataset and the cadaver lung dataset, AirwayNet outperforms BifurcationNet ($p < 0.05$). The difference in performance is likely due to two main causes. First, BifurcationNet's particle filter is sensitive to estimates from previous frames, so incorrectly classifying airways makes future predictions more difficult. AirwayNet, on the other hand, does not require a particle filter and does not factor previous estimates into the current estimate, so mistakes are limited to the specific frame. Second, to classify an airway, AirwayNet can leverage all the information in image space, while BifurcationNet uses only 7 airway characteristics plus insertion and state history. The image space information evidently makes up for the seemingly harder task of classifying airways without prior context.

Closer inspection of the causes of failure for BifurcationNet reveals that when the CNN sees nonexistent bifurcations or airways for several consecutive frames, the filter becomes increasingly confident about an incorrect belief of its location, from which it cannot recover. AirwayNet would similarly occasionally see nonexistent bifurcations and airways, but its failures were primarily due to misclassifying visible bifurcations and airways.

In both datasets, the wide range of performance suggests a range in difficulty of the localization task between lungs and sequences. Several factors may contribute to the difficulty of the task. We identify three key factors: the image domain gap, lung geometry differences, and sequence-specific challenges.

The image domain gap refers to the similarity between the holdout lung's camera images, $I_{x \in L_j}^{cam}$, and the training images. For example, certain cadaver lungs contain artifacts that don't exist in the other lungs, shown in Fig. 2. When trained on images without similar artifacts, both networks make inaccurate estimates. One approach to overcome this challenge is to train on enough lungs to cover the possible set of image domains. As shown in Fig. 5, training on more lungs tends to improve performance for both algorithms. Extrapolating this trend to many more training lungs would likely improve that performance; however, it is unclear if that would be sufficient to reach the goal of consistently accurate airway classification across all the lungs in this dataset. The GAN was designed to help bridge the domain gap between simulated and camera domains, $I^{rsim}, I^{cam}$. When tested on $I_{x \in L_j}^{gan}$ images, BifurcationNet trained on $I^{rsim}$ showed improved CNN results compared being tested to $I_{x \in L_j}^{cam}$, shown in Table S2, but the effect was too small to change the aggregate localization results.

The geometry differences between lungs, including the position and orientation of the airways, seem to influence the algorithms' performance. One of the most surprising results was that training AirwayNet on simulated images outperformed the models trained on camera images (p<0.01). This suggests the benefit of training on the lung-specific geometry outweighs the benefit of training on up to 10 different lung geometries; however, this result is not independent from the image domain gap. As shown in (*23*), domain randomization helps bridge the domain gap between the camera images and the simulated images.

Finally, within the same lung, different sequences present different challenges to the networks. For example, Fig. 4 shows the performance in the lung phantom varies between sequences despite a uniform image domain and a static geometry. This challenge affected BifurcationNet more profoundly than AirwayNet, as BifurcationNet failed to track half the sequences. In these sequences, the particle filter struggled to differentiate between the possible airways based on the bifurcation location and airway orientation. Approaching the same airways from a different perspective enabled the network to correctly classify the airways.

To overcome the challenges in these datasets and improve the algorithms, we suggest three approaches: labeling more human lung data, improving the simulation of the bronchoscopic images, and extracting more information from the images. More labeled human data would serve as both training and test data for proposed algorithms. Improving the labeling process would assist this aim as this manual and time-consuming approach will not easily scale to tens or hundreds more lungs. High quality labels are critical for training and especially critical for test sets. Next, improving the simulation tools would improve the performance of networks trained on simulated images. This may involve taking higher resolution CT scans, improving the rendering framework and domain randomization capabilities, or training better image-style transfer techniques like the GAN presented here. For the algorithms themselves, it appears that more information can be extracted from the images to help classify airways, as shown by AirwayNet's superiority to BifurcationNet. Adding information beyond what's present in the images is another opportunity for improvement, especially because the image quality is variable across lungs. For instance, emphysema can cause occlusions to the camera view. Incorporating EM sensors and other sensing modalities with these image-based localization methods would improve their performance.

By advancing the localization performance of image-based algorithms, the field can improve the feedback used by physicians. This is particularly relevant for robotic bronchoscopy, where a physician drives the bronchoscope using a controller. The localization results directly integrate with the robot telemetry to give turn-by-turn navigation and augment the raw camera feedback with 3D renderings of the present location in the patient's CT. Further down the road, assisted driving, or even fully automated closed-loop control of robotic bronchoscopes is possible.

Vision Benchmark Suite," (available at www.cvlibs.net/datasets/kitti).

**Acknowledgments: Funding:** This work was supported by Auris Health, Inc., the National Institutes of Health (NIH) NIGMS Training Grant in Biotechnology (5T32GM008412), the Siebel Scholarship, and the NVIDIA GPU Grant.  **Author contributions:** Jake Sganga contributed to conceptualization, data curation, analysis, funding acquisition, algorithm design, writing the original draft, making the figures. David Eng contributed to conceptualization, data curation, analysis, editing the paper, and making figures. Chauncey Graetzel contributed to funding acquisition, data curation, editing the paper. David Camarillo contributed to funding acquisition, editing the paper, and supervising the research. **Competing interests:** David Eng and David Camarillo declare no competing interests. Chauncey Graetzel is an employee of Auris Health, Inc. and Jake Sganga is a consultant to Auris Health, Inc.

**Supplementary Materials:**

Materials and Methods

Figures S1-S3

Tables S1-S4

References (*24–33*)

# Supplementary Materials for

## Deep Learning for Localization in the Lung

Jake Sganga, David Eng, Chauncey Graetzel, David B. Camarillo.

Correspondence to: sganga@stanford.edu

**This PDF file includes:**

## Materials and Methods

### Localization Control Loop

Shown in Fig. S1 and S2, at every step in the localization task, an image from the bronchoscope, $I_{x_t}^{cam}$, at time $t$ from the position, $x_t$, and the current absolute robot insertion (mm), $i_t$, is provided to the localization algorithm and the algorithm outputs a 6 degree of freedom (DOF) location estimate in CT frame, $\hat{x}_t$, along with the set of visible airways and their positions and orientations, $a_t$, with respect to camera frame. In the case of localization algorithms that only provide a 6 DOF location estimate, the location estimate, $\hat{x}_t$, can be used to determine the visible airways based on the camera field of view and lies within a max visible distance (set to 3 cm) and the preoperative CT scan.

Both networks identify a set of common characteristics for the visible airways in each image. Each airway is characterized through 2 visibility measures, *isVis* if the airway is visible and *hasVisChild* if the airway's bifurcation is visible. For each airway, the networks regress the camera frame position $(x, y, z)$ of the furthest point on the airway and its angle $(\alpha, \beta)$. BifurcationNet outputs the airway characteristics of up to four visible airways into a novel particle filter that classifies the airways based on the most probable airways in the CT map, shown in Fig. 6. AirwayNet, on the other hand, classifies each visible airway as an airway in the CT map, denoted by "Airway ID" in Fig. S2. By matching the airways directly to the CT map, it does not need a particle filter. When AirwayNet is trained on simulated data from the holdout lung, $I_{x \in L_{i=j}}^{rsim}$, it can be trained on every airway in the lung CT. When training integrates information from multiple lungs, only 31 conserved airways (up to the 5th generation) are used for training. Both algorithms use the airway assignments to estimate the bronchoscope location in the lung CT in real-time. If an airway's bifurcation is visible (*hasVisChild*) and two of the airway's children are also visible (*isVis*), $\hat{x}_t$ is backed out from the parent airway's position and angle and the angles of the children airways.

<u>Convolutional Neural Networks</u>

Both BifurcationNet and AirwayNet consist of deep residual convolutional networks (CNN) and a single fully connected layer, which produces the output of the corresponding dimensions for each algorithm. The residual parts of our network implement the 18-layer architecture with an embedding length of 128, described in He et al. (*27*). The CNN was implemented in Tensorflow, version 1.9 (*28*).

<u>Training</u>

The networks are trained using Adam optimization to minimize a weighted L2 loss function. The loss function combines the sigmoid cross entropy loss of the two boolean outputs (*isVis*, *hasVisChil*d) and a mean L2 loss on airway position $l(x_{airway}, \hat{x}_{airway})$ and airway angle $l(\alpha_{airway}, \hat{\alpha}_{airway})$. To relate position and rotation errors, we chose a 1 mm:5.7° ratio, which roughly relates to the fact that a 5.7° $e_d$ angle error results in an error of 1 mm for a location 10 mm in front of the camera. Both networks weighted the combined loss by the distance from the camera of the airway ($6 - 0.2 \times depth$) to focus the network's attention to nearby airways. Additionally, the relative weighting between the classifications and the regressions were varied, but not thoroughly examined. The relative weighting used for the training of the networks presented in this work are as follows - BifurcationNet: 1x classification, 1x airway pose, additional 2x airway pose when *hasVisChild*=True, 10x airway angle; AirwayNet: 2x classification, 1x airway pose, 10x airway angle.

For the lung phantom experiment in Fig. 3, both models were trained for 60k steps. Both models were trained with Adam optimization and with a learning rate decay of 0.75. For the cadaver lung experiments in Fig. 4 and Fig. 5, all models were trained for 30k steps. All models were trained with Adam optimization and with a learning rate decay of 0.9.

All models were trained on an NVIDIA Titan X GPU.

Datasets

In all experiments, $I^{rsim}$ were randomly offset in translation, view direction and roll from the recorded image locations in the lung with standard deviation 2 mm, 11°, 11°, respectively. Tobin et al. first introduced $I^{rsim}$ images, which were rendered with randomized parameters to train their CNN to be indifferent to these changes (*24*). We randomized each rendering parameter with a normal distribution centered about the default rendering parameters. For these experiments, brightness, attenuation factor, specular intensity, and ambient intensity were all varied by 1, 0.001, 0.1, and 0.1, respectively. The rendering framework used to generate the simulated images was developed in PyOpenGL, and is the same framework described in (*23*). The rendering parameters are based on Higgins et al. with a field of view of 60° (*31*). Images are rendered at 60 Hz on a PC with no accelerations.

$I^{cam}$ were randomly rolled with standard deviation 14°. In addition, all images ($I^{cam}$, $I^{rsim}$) were augmented with randomized Gaussian smoothing, and on half of the images, we added independent per-pixel noise and white noise occlusions of various sizes. Finally, all images were grayscaled and normalized per-image to 0 mean and unit scaling.

For the lung phantom experiment in Fig. 3, 995k $I^{rsim}$ were used for BifurcationNet and 200k $I^{rsim}$ were used for AirwayNet. In all cadaver lung experiments, only 10k examples per dataset type were used to limit the time needed to train each model. All of the datasets contained randomly mixed examples evenly distributed from the sampled input lungs.

For the cadaver lung experiment in Fig. 4, 68 representative sequences from 11 cadaver lungs were collected for testing. To test how training a model with data from a single cadaver lung performed on the sequences in a holdout cadaver lung, we trained 5 models, each on $I^{cam}$ from a randomly selected cadaver lung (without replacement), per holdout cadaver lung. To test how training a model with data from five cadaver lungs performed on the sequences in a holdout cadaver lung, we trained 3 models, each on $I^{cam}$ from a set of 5 randomly-sampled cadaver lungs, per holdout cadaver lung. To test how training a model with data from 10 cadaver lungs performed on the sequences in the only holdout cadaver lung, we trained 1 model on $I^{cam}$ from all 10 of the available cadaver lungs, per holdout cadaver lung.

<u>GAN</u>

To mitigate the error incurred by training a model on $I^{rsim}$ and testing it on $I^{cam}$ in the cadaver lungs, we implemented Cycle GAN to learn the transformation from $I^{cam}$ in the cadavers to $I^{sim}$. We trained it as described in (*25*), with an additional loss term that imposed a penalty for pixel-wise differences between the generated and true simulated images.

To construct the dataset to train our GANs, we split our ($I^{cam}$, $I^{sim}$) pairs into two groups, such that each group contained image pairs from the same number of cadaver lungs and image pairs corresponding to the same cadaver lungs were assigned to the same group. We trained two GAN's in total, one on each of the groups. For the cadaver lung experiment in Fig. 4 involving the GAN, the GAN that was trained on data from the first group of cadaver lungs was applied to images from the second group of cadaver lungs during testing and vice versa. To train a GAN, we sampled 2.5k image pairs per group. We alternated training the generator and discriminator of the GAN as described in the literature, with slight modification. On each iteration, we continued training the generator until it deceived the discriminator at a rate of 0.5. All components of the GAN were trained with Adam optimization.

<u>Particle</u> <u>Filter</u> <u>for</u> <u>BifurcationNet</u>

Because the output of the BifurcationNet CNN is generic to any airway in the lung skeleton, a particle filter is needed to pair the visible airways with the airways in the underlying CT. A novel particle was designed to solve this identification challenge. When a bifurcation is visible and at least two children airways are visible, the particle filter determines the probability the measurement fits a given CT bifurcation. The particle filter also calculates the prior probability of seeing that CT bifurcation. With the measurement probability and the prior probability, the filter calculates the posterior probability of the bronchoscope seeing that bifurcation. In the work shown here, three bifurcations with the highest prior probability were compared. The filter assigns the bifurcation with the maximum posterior probability to the visible airways and uses that assignment to calculate the estimated bronchoscope position, $\hat{x}_t$. Each step is detailed below:

*Measurement Probability*: the probability an observation matches a given bifurcation in the CT is based on how well the children airways align with the expected airways once the bifurcation point and the parent airway direction is aligned with the CT. To resolve the roll about the parent axis, the different child airway assignments to the underlying CT bifurcation were permuted and compared. For a given airway assignment, the optimal roll about the parent axis was calculated by minimizing the weighted average of the airway angle offsets. The probability of the fit is calculated based on a Gaussian distribution over the cosine angle difference between the measured child airway directions and CT airway directions.

$$Let \ p(x|\mu,\sigma) = \ \frac{1}{\sqrt{2\pi\sigma^2}}e^{-\frac{(x-\mu)^2}{2\sigma^2}}$$

$$prob_{fit} = \sum_i(p(\arccos(a_i^T\hat{a}_i)|0,\sigma_{fit})) \ / \ n$$

*Prior*: the bifurcation prior was calculated based on the current robot insertion length and the filter's previous state. The insertion probability, $prob_{ins}$, is based on the distance between a given CT bifurcation length from the trachea, $z_{bif}$, and the robot insertion, $i_t$, plus the observed bifurcation depth in camera frame, $\hat{z}_{bif}$. The prior airway probability, $prob_{airways}$, increases the probability of bifurcations near previously visible airways. If the bifurcation is 1, 2, or 3 generations removed from a visible airway, $d(a_j, b_i)$, its relative probability increments by 1, 0.1 and 0.01. For $prob_x$, the previously estimated 3DOF location, $\hat{x}_{t-1}$, is used to calculate a 3D multivariate Gaussian with the particle estimate. Additionally, a roll probability is included to prevent sudden rotations and the previous roll value.

$$prob_{bif} = prob_{ins} * prob_{airways} * prob_x$$

$$prob_{ins} = p\big(i_t + \hat{z}_{bif} - z_{bif} | 0, \sigma_{ins}\big)$$

$$prob_{airways} = \frac{\sum_j p_{gen}(a_{t-1}[j], b_i)}{\sum_i \sum_j p_{gen}(a_{t-1}[j], b_i)}; p_{gen}(a_j, b_i) = \begin{cases} 1 \ if \ d(a_j, b_i) = 1 \\ 0.1 \ if \ d(a_j, b_i) = 2 \\ 0.01 \ if \ d(a_j, b_i) = 3 \end{cases}$$

$$prob_x = p(\hat{x}_{t-1} - \tilde{x}_i | 0, \Sigma_x)$$

$$prob_{roll} = p(\hat{r}_{t-1} - \tilde{r}_i | 0, \sigma_{roll})$$

Probability for bifurcation $i$:

$$prob_{\tilde{x}_i} = prob_{fit}(i) * prob_{bif}(i) * prob_{roll}(i)$$

$$\hat{x}_t = \max_i prob_{\tilde{x}_i}$$

Figure S3 shows the resulting change in the average F1 score for a single sequence in the lung phantom when each of the hard-coded variances for these probability measures are changed. The most important inputs to the system are the insertion variance, fit variance, and roll variance. This set of filter parameters was found through manual optimization on a run in the lung phantom sequence and was used for all cadaver lung experiments.

Localization Evaluation

We chose to provide several metrics that cover two aspects of navigation - the identification of visible airways and the tracking error, $(e_p, e_d, e_r)$. The visible airways are critical for navigation decisions, and the location-based metrics are important for determining the proximity to walls and the final biopsy target.

We define a visible airway, *isVis*, as one whose centerline lies within 3 cm of the bronchoscope location and would lie within the camera's field of view. An airway has visible children, *hasVisChild*, if the airway's bifurcation satisfies the same visibility criteria. To compare the estimated visible airways to the labeled airways, we look at precision and recall, which focus on the true values (visible airways), as there are many more invisible airways at any given moment than visible ones. To determine the precision and recall on the visible airways, let $a_t$ be the set of airways visible at $x_t$, and $\hat{a}_t$ be the set of airways estimated to be visible. Recall is defined as the true positive rate, $\sum a_t \cap \hat{a}_t / \sum a_t$, while precision is the positive predictive value, $\sum a_t \cap \hat{a}_t / \sum \hat{a}_t$. The F1 score is the harmonic mean of the precision and recall, $2 * recall * precision / (recall + precision)$.

For the tracking errors, $(e_p, e_d, e_r)$, we reference Merritt et al (*19*).

<u>*Labeling Datasets*</u>

All $I^{cam}$ images were recorded by manually driving a robotic bronchoscope (Monarch Platform, Auris Health Inc.). While driving, a 6-DOF electromagnetic sensor (Northern Digital Inc.) tracked the bronchoscope and provided an initial, coarse estimate of the location in CT frame $\tilde{x}_t$ from which each $I^{cam}_{x_t}$ was captured. However, a single rigid registration of the sensor's output resulted in errors greater than several millimeters, so local registrations were iteratively refined to improve the label quality. For the lung phantom dataset, local registrations were performed by manually translating and rotating nearby points (within a 2 cm cube) until the pixel-wise error $\left\|I_x^{cam} - I_x^{sim}\right\|_2$ reached a local minimum. Powell's method was used in conjunction with manual optimization to refine $\hat{x}_t$ to a ground-truth location in CT frame $x_t$ (*29*). In the cadaver lung sequences, the pixel-wise error led to errors greater than several millimeters in registration, so a purely manual approach was used. As such, its quality was subjective to the operator.

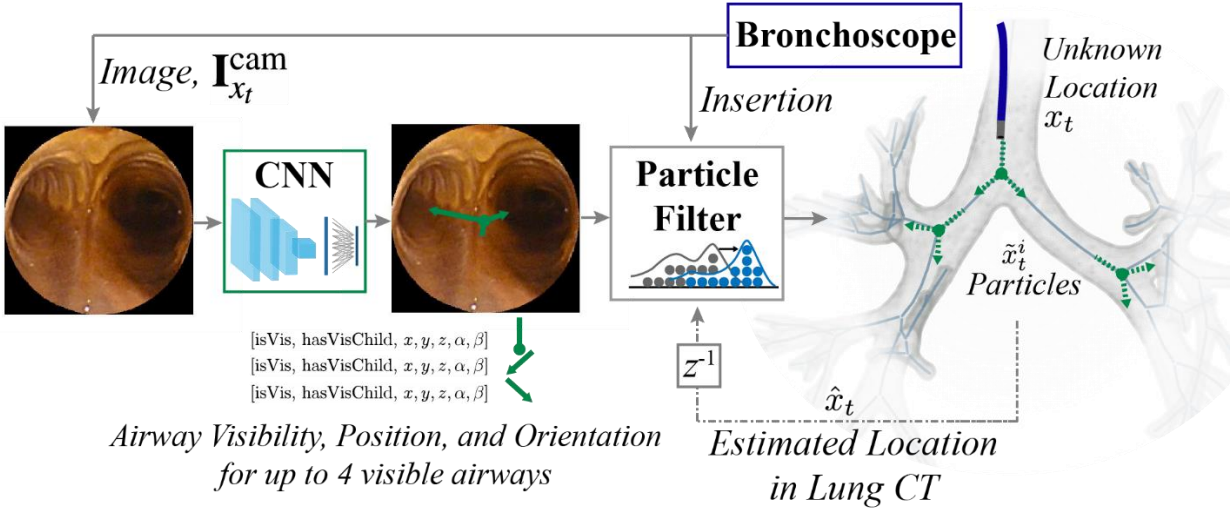

*Airway Visibility, Position, and Orientation
for up to 4 visible airways*

*Estimated Location
in Lung CT*

**Fig. S1.**

The control loop for BifurcationNet is shown. A camera image from the bronchoscope's true position, $I_{x_t}^{cam}$, at time $t$ passes through a trained CNN (ResNet-18) and outputs a matrix, representing the characteristics of 4 airways. Two booleans capture the airway's visibility and the airway's bifurcation visibility. The CNN also outputs the camera frame position $(x, y, z)$ in mm of the furthest visible point of the airway, which corresponds to the bifurcation when it is visible, and $(\alpha, \beta)$ representing the XYZ Euler angles for camera frame angle about X and Y, respectively (26). The 4 airways are ordered based on the position proximity to the camera and angle of the airway. The particle filter compares how this measurement relates to the most likely bifurcations based on the previous state and the current insertion from the robot, finding the posterior probability of each particle. The bifurcation with the highest posterior probability is selected and used to calculate $\hat{x}_t$. This is fed into the particle filter on the next time step.

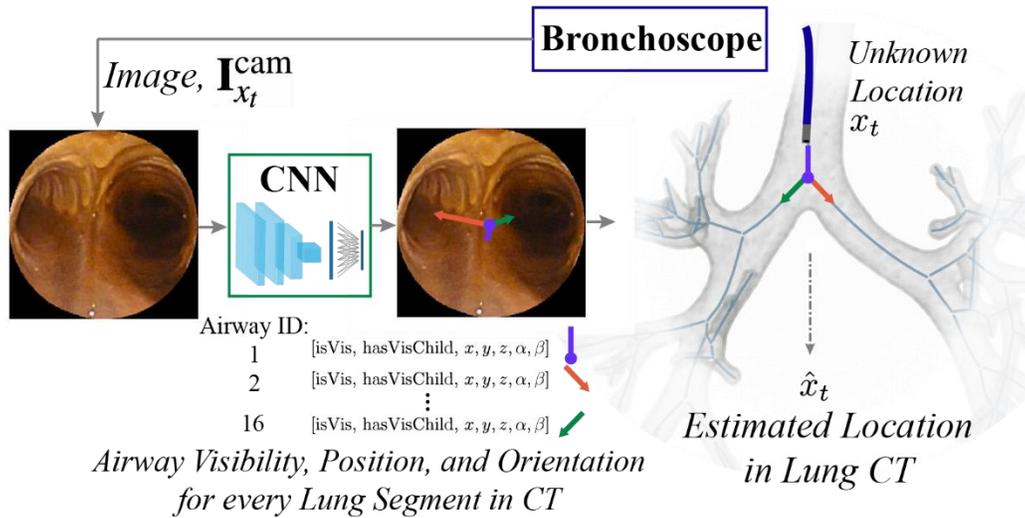

**Fig. S2.**

The control loop for AirwayNet is shown. Similar to BifurcationNet, a camera image from the bronchoscope's true position, $I_{x_t}^{cam}$, passes through a trained CNN (ResNet-18) and outputs a matrix. In this case, the matrix represents the airway characteristics for every airway in the lung skeleton. Identifying which airway is visible is a classification task for this model rather than a task for the particle filter. If the identified airway *hasVisChild* and the children airways *isVis* and consistent with the lung skeleton, the position $\hat{x}_t$ is backed out from the measurement.

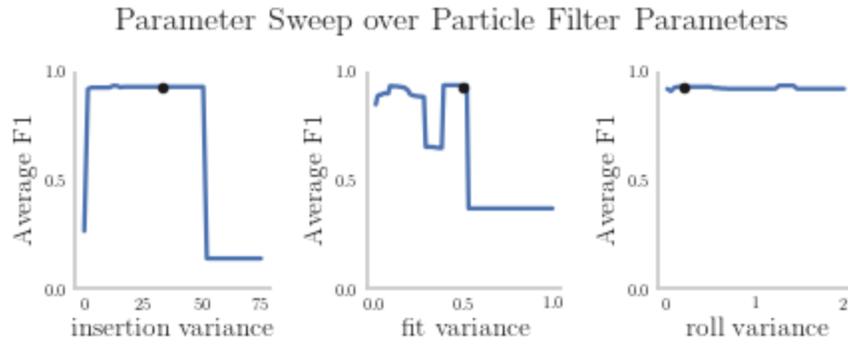

**Fig. S3.**

In each subplot, the specified parameter in the BifucationNet particle filter was varied while keeping all other parameters constant. For each value, BifurcationNet was evaluated on a sequence in the lung phantom, and the average F1 score of visible airways in the run was recorded. The black dot represents the value of the parameter that was used in all BifurcationNet plots in this paper.

**Table S1.**

The performance of BifurcationNet and AirwayNet in lung phantom, cadaver lungs, and for a single cadaver lung, also shown in Fig. 3, 4, and 5, respectively.

| Type | Train set | Test set | F1 Score by Airway Generation | | | | | Tracking Error | | |
|---|---|---|---|---|---|---|---|---|---|---|
| | | | 1 | 2 | 3 | 4 | 5 | $e_p$ (mm) | $e_d$ (°) | $e_r$ (°) |
| *Performance for BifurcationNet and AirwayNet over 13 sequences in 1 lung phantom* | | | | | | | | | | |
| **BifurcationNet** | $I^{rsim}_{x_t \in L_{i=j}}$ | $I^{cam}_{x_t \in L_j}$ | 0.983 | 0.898 | 0.766 | 0.316 | 0.383 | 3.1 | 6.2 | 9.5 |
| **AirwayNet** | $I^{rsim}_{x_t \in L_{i=j}}$ | $I^{cam}_{x_t \in L_j}$ | 0.994 | 0.908 | 0.957 | 0.888 | 0.878 | 10.5 | 9.3 | 8.7 |
| *Performance for BifurcationNet and AirwayNet over 68 sequences in 11 cadavers* | | | | | | | | | | |
| **BifurcationNet** | $I^{cam}_{x_t \in L_{i \neq j} \times 1}$ | $I^{cam}_{x_t \in L_j}$ | 0.883 | 0.262 | 0.079 | 0.041 | 0.031 | 8.8 | 21.2 | 13.4 |
| **BifurcationNet** | $I^{cam}_{x_t \in L_{i \neq j} \times 5}$ | $I^{cam}_{x_t \in L_j}$ | 0.891 | 0.293 | 0.053 | 0.045 | 0.032 | 8.4 | 19.5 | 14.3 |
| **BifurcationNet** | $I^{cam}_{x_t \in L_{i \neq j} \times 10}$ | $I^{cam}_{x_t \in L_j}$ | 0.904 | 0.354 | 0.106 | 0.064 | 0.118 | 8.0 | 20.3 | 16.1 |
| **BifurcationNet** | $I^{rsim}_{x_t \in L_{i=j}}$ | $I^{cam}_{x_t \in L_j}$ | 0.921 | 0.482 | 0.203 | 0.162 | 0.014 | 6.3 | 15.9 | 16.1 |
| **BifurcationNet** | $I^{rsim}_{x_t \in L_{i=j}}$ | $I^{gan}_{x_t \in L_j}$ | 0.914 | 0.443 | 0.116 | 0.046 | 0.031 | 6.2 | 16.9 | 18.4 |
| **AirwayNet** | $I^{cam}_{x_t \in L_{i \neq j} \times 1}$ | $I^{cam}_{x_t \in L_j}$ | 0.391 | 0.330 | 0.103 | 0.030 | 0.001 | 15.6 | 32.8 | 60.4 |
| **AirwayNet** | $I^{cam}_{x_t \in L_{i \neq j} \times 5}$ | $I^{cam}_{x_t \in L_j}$ | 0.624 | 0.482 | 0.088 | 0.006 | 0.000 | 11.5 | 27.2 | 46.6 |
| **AirwayNet** | $I^{cam}_{x_t \in L_{i \neq j} \times 10}$ | $I^{cam}_{x_t \in L_j}$ | 0.639 | 0.586 | 0.109 | 0.001 | 0.000 | 11.1 | 25.0 | 44.2 |
| **AirwayNet** | $I^{rsim}_{x_t \in L_{i=j}}$ | $I^{cam}_{x_t \in L_j}$ | 0.895 | 0.651 | 0.449 | 0.318 | 0.078 | 7.3 | 11.1 | 9.5 |
| **AirwayNet** | $I^{rsim}_{x_t \in L_{i=j}}$ | $I^{gan}_{x_t \in L_j}$ | 0.580 | 0.510 | 0.543 | 0.456 | 0.077 | 8.6 | 14.4 | 10.4 |
| *Performance for BifurcationNet and AirwayNet over 1 sequence in 1 cadaver* | | | | | | | | | | |
| **BifurcationNet** | $I^{rsim}_{x_t \in L_{i=j}}$ | $I^{cam}_{x_t \in L_j}$ | 0.869 | 0.989 | 0.638 | 0.756 | 0.333 | 4.8 | 12.4 | 6.2 |
| **AirwayNet** | $I^{rsim}_{x_t \in L_{i=j}}$ | $I^{cam}_{x_t \in L_j}$ | 0.887 | 0.964 | 0.970 | 0.916 | 0.500 | 3.6 | 8.9 | 9.1 |

**Table S2.**

The F1 score of the *isVis* CNN output for BifurcationNet and AirwayNet in lung phantom, cadaver lungs, and for a single cadaver lung, also shown in Fig. 3, 4, and 5, respectively.

| Type | Train set | Test set | F1 Score by Airway Generation | | | | |
|---|---|---|---|---|---|---|---|
| | | | 1 | 2 | 3 | 4 | 5 |
| *CNN performance over 13 sequences in 1 lung phantom* | | | | | | | |
| **BifurcationNet** | $I^{\text{rsim}}_{x_t \in L_{i=j}}$ | $I^{\text{cam}}_{x_t \in L_j}$ | 0.984 | 0.938 | 0.964 | 0.995 | 0.969 |
| **AirwayNet** | $I^{\text{rsim}}_{x_t \in L_{i=j}}$ | $I^{\text{cam}}_{x_t \in L_j}$ | 0.994 | 0.917 | 0.937 | 0.880 | 0.862 |
| *CNN performance over 68 sequences in 11 cadavers* | | | | | | | |
| **BifurcationNet** | $I^{\text{cam}}_{x_t \in L_{i \neq j} \times 1}$ | $I^{\text{cam}}_{x_t \in L_j}$ | 0.912 | 0.840 | 0.879 | 0.953 | 0.925 |
| **BifurcationNet** | $I^{\text{cam}}_{x_t \in L_{i \neq j} \times 5}$ | $I^{\text{cam}}_{x_t \in L_j}$ | 0.912 | 0.878 | 0.880 | 0.955 | 0.916 |
| **BifurcationNet** | $I^{\text{cam}}_{x_t \in L_{i \neq j} \times 10}$ | $I^{\text{cam}}_{x_t \in L_j}$ | 0.923 | 0.897 | 0.872 | 0.966 | 0.881 |
| **BifurcationNet** | $I^{\text{rsim}}_{x_t \in L_{i=j}}$ | $I^{\text{cam}}_{x_t \in L_j}$ | 0.933 | 0.888 | 0.881 | 0.926 | 0.892 |
| **BifurcationNet** | $I^{\text{rsim}}_{x_t \in L_{i=j}}$ | $I^{\text{gan}}_{x_t \in L_j}$ | 0.935 | 0.886 | 0.929 | 0.920 | 0.921 |
| **AirwayNet** | $I^{\text{cam}}_{x_t \in L_{i \neq j} \times 1}$ | $I^{\text{cam}}_{x_t \in L_j}$ | 0.382 | 0.419 | 0.129 | 0.028 | 0.002 |
| **AirwayNet** | $I^{\text{cam}}_{x_t \in L_{i \neq j} \times 5}$ | $I^{\text{cam}}_{x_t \in L_j}$ | 0.615 | 0.594 | 0.112 | 0.006 | 0.000 |
| **AirwayNet** | $I^{\text{cam}}_{x_t \in L_{i \neq j} \times 10}$ | $I^{\text{cam}}_{x_t \in L_j}$ | 0.634 | 0.678 | 0.167 | 0.001 | 0.000 |
| **AirwayNet** | $I^{\text{rsim}}_{x_t \in L_{i=j}}$ | $I^{\text{cam}}_{x_t \in L_j}$ | 0.886 | 0.705 | 0.450 | 0.313 | 0.196 |
| **AirwayNet** | $I^{\text{rsim}}_{x_t \in L_{i=j}}$ | $I^{\text{gan}}_{x_t \in L_j}$ | 0.574 | 0.619 | 0.561 | 0.484 | 0.206 |
| *CNN performance over 1 sequence in 1 cadaver* | | | | | | | |
| **BifurcationNet** | $I^{\text{rsim}}_{x_t \in L_{i=j}}$ | $I^{\text{cam}}_{x_t \in L_j}$ | 0.869 | 0.941 | 0.971 | 1.00 | 1.00 |
| **AirwayNet** | $I^{\text{rsim}}_{x_t \in L_{i=j}}$ | $I^{\text{cam}}_{x_t \in L_j}$ | 0.887 | 0.947 | 0.882 | 0.916 | 1.00 |

**Table S3.**

The F1 score of the *hasVisChild* CNN output for BifurcationNet and AirwayNet in lung phantom, cadaver lungs, and for a single cadaver lung, also shown in Fig. 3, 4, and 5, respectively.

| Type | Train set | Test set | F1 Score by Airway Generation | | | | |
|---|---|---|---|---|---|---|---|
| | | | 1 | 2 | 3 | 4 | 5 |
| *CNN performance over 13 sequences in 1 lung phantom* | | | | | | | |
| **BifurcationNet** | $I^{rsim}_{x_t \in L_{i=j}}$ | $I^{cam}_{x_t \in L_j}$ | 0.997 | 0.873 | 0.913 | 0.909 | 0.902 |
| **AirwayNet** | $I^{rsim}_{x_t \in L_{i=j}}$ | $I^{cam}_{x_t \in L_j}$ | 0.999 | 0.818 | 0.882 | 0.881 | 0.886 |
| *CNN performance over 68 sequences in 11 cadavers* | | | | | | | |
| **BifurcationNet** | $I^{cam}_{x_t \in L_{i \neq j} \times 1}$ | $I^{cam}_{x_t \in L_j}$ | 0.741 | 0.584 | 0.532 | 0.686 | 0.599 |
| **BifurcationNet** | $I^{cam}_{x_t \in L_{i \neq j} \times 5}$ | $I^{cam}_{x_t \in L_j}$ | 0.799 | 0.640 | 0.528 | 0.667 | 0.631 |
| **BifurcationNet** | $I^{cam}_{x_t \in L_{i \neq j} \times 10}$ | $I^{cam}_{x_t \in L_j}$ | 0.835 | 0.675 | 0.558 | 0.703 | 0.494 |
| **BifurcationNet** | $I^{rsim}_{x_t \in L_{i=j}}$ | $I^{cam}_{x_t \in L_j}$ | 0.869 | 0.689 | 0.550 | 0.632 | 0.606 |
| **BifurcationNet** | $I^{rsim}_{x_t \in L_{i=j}}$ | $I^{gan}_{x_t \in L_j}$ | 0.848 | 0.702 | 0.664 | 0.740 | 0.706 |
| **AirwayNet** | $I^{cam}_{x_t \in L_{i \neq j} \times 1}$ | $I^{cam}_{x_t \in L_j}$ | 0.197 | 0.170 | 0.038 | 0.019 | 0.003 |
| **AirwayNet** | $I^{cam}_{x_t \in L_{i \neq j} \times 5}$ | $I^{cam}_{x_t \in L_j}$ | 0.452 | 0.277 | 0.026 | 0.003 | 0.000 |
| **AirwayNet** | $I^{cam}_{x_t \in L_{i \neq j} \times 10}$ | $I^{cam}_{x_t \in L_j}$ | 0.464 | 0.408 | 0.045 | 0.001 | 0.000 |
| **AirwayNet** | $I^{rsim}_{x_t \in L_{i=j}}$ | $I^{cam}_{x_t \in L_j}$ | 0.810 | 0.456 | 0.372 | 0.271 | 0.222 |
| **AirwayNet** | $I^{rsim}_{x_t \in L_{i=j}}$ | $I^{gan}_{x_t \in L_j}$ | 0.450 | 0.411 | 0.484 | 0.467 | 0.253 |
| *CNN performance over 1 sequence in 1 cadaver* | | | | | | | |
| **BifurcationNet** | $I^{rsim}_{x_t \in L_{i=j}}$ | $I^{cam}_{x_t \in L_j}$ | 0.857 | 0.713 | 0.703 | 0.810 | 0.667 |
| **AirwayNet** | $I^{rsim}_{x_t \in L_{i=j}}$ | $I^{cam}_{x_t \in L_j}$ | 0.820 | 0.847 | 0.935 | 0.766 | 1.00 |

**Table S4.**

The average position loss, $e_p$, on the airway positions and the average direction loss, $e_d$, on the airway directions for the CNN output for BifurcationNet and AirwayNet in lung phantom, cadavers, and for a single cadaver, also shown in Fig. 3, 4, and 5, respectively

| | | | Tracking Error | |
|---|---|---|---|---|
| **Type** | **Train set** | **Test set** | $e_p$ (mm) | $e_d$ (°) |
| *CNN performance over 13 sequences in 1 lung phantom* | | | | |
| **BifurcationNet** | $I^{rsim}_{x_t \in L_{i=j}}$ | $I^{cam}_{x_t \in L_j}$ | 2.8 | 9.3 |
| **AirwayNet** | $I^{rsim}_{x_t \in L_{i=j}}$ | $I^{cam}_{x_t \in L_j}$ | 3.2 | 6.8 |
| *CNN performance over 68 sequences in 11 cadavers* | | | | |
| **BifurcationNet** | $I^{cam}_{x_t \in L_{i \neq j} \times 1}$ | $I^{cam}_{x_t \in L_j}$ | 8.4 | 31.0 |
| **BifurcationNet** | $I^{cam}_{x_t \in L_{i \neq j} \times 5}$ | $I^{cam}_{x_t \in L_j}$ | 8.0 | 30.0 |
| **BifurcationNet** | $I^{cam}_{x_t \in L_{i \neq j} \times 10}$ | $I^{cam}_{x_t \in L_j}$ | 7.7 | 29.2 |
| **BifurcationNet** | $I^{rsim}_{x_t \in L_{i=j}}$ | $I^{cam}_{x_t \in L_j}$ | 6.4 | 22.9 |
| **BifurcationNet** | $I^{rsim}_{x_t \in L_{i=j}}$ | $I^{gan}_{x_t \in L_j}$ | 5.6 | 21.2 |
| **AirwayNet** | $I^{cam}_{x_t \in L_{i \neq j} \times 1}$ | $I^{cam}_{x_t \in L_j}$ | 11.7 | 36.3 |
| **AirwayNet** | $I^{cam}_{x_t \in L_{i \neq j} \times 5}$ | $I^{cam}_{x_t \in L_j}$ | 10.0 | 34.1 |
| **AirwayNet** | $I^{cam}_{x_t \in L_{i \neq j} \times 10}$ | $I^{cam}_{x_t \in L_j}$ | 9.2 | 32.9 |
| **AirwayNet** | $I^{rsim}_{x_t \in L_{i=j}}$ | $I^{cam}_{x_t \in L_j}$ | 5.7 | 14.9 |
| **AirwayNet** | $I^{rsim}_{x_t \in L_{i=j}}$ | $I^{gan}_{x_t \in L_j}$ | 5.4 | 15.0 |
| *CNN performance over 1 sequence in 1 cadaver* | | | | |
| **BifurcationNet** | $I^{rsim}_{x_t \in L_{i=j}}$ | $I^{cam}_{x_t \in L_j}$ | 5.4 | 26.0 |
| **AirwayNet** | $I^{rsim}_{x_t \in L_{i=j}}$ | $I^{cam}_{x_t \in L_j}$ | 2.6 | 10.7 |